\documentclass{article}
\usepackage{spconf,amsmath,epsfig}
\usepackage{subfigure,graphicx}
\usepackage{multirow}
\usepackage{color}

\begin{document}\sloppy

% Example definitions.
% --------------------
\def\x{{\mathbf x}}
\def\L{{\cal L}}

% Title.
% ------
\title{MILD: MULTI-INDEX HASHING FOR APPEARANCE BASED	\\ LOOP CLOSURE DETECTION}
%
% Single address.
% ---------------
\name{Lei Han, Lu Fang}

\address{Robotic Institute, Hong Kong University of Science and Technology \\ lhanaf@connect.ust.hk, eefang@ust.hk}

\maketitle

\begin{abstract}
Loop Closure Detection (LCD) has been proved to be extremely useful in global consistent visual Simultaneously Localization and Mapping (SLAM) and appearance-based robot relocalization. Methods exploiting binary features in bag of words representation have recently gained a lot of popularity for their efficiency, but suffer from low recall due to the inherent drawback that high dimensional binary feature descriptors lack well-defined centroids. In this paper, we propose a realtime LCD approach called MILD (Multi-Index Hashing for Loop closure Detection), in which image similarity is measured by feature matching directly to achieve high recall without introducing extra computational complexity with the aid of Multi-Index Hashing (MIH). A theoretical analysis of the approximate image similarity measurement using MIH is presented, which reveals the trade-off between efficiency and accuracy from a probabilistic perspective. Extensive comparisons with state-of-the-art LCD methods demonstrate the superiority of MILD in both efficiency and accuracy.
\end{abstract}
\section{Introduction}
\label{sec:intro}
Visual Loop Closure Detection (LCD) tries to detect previously visited places based on appearance information of the scene. LCD can play an important part in Global Consistent Visual Simultaneous Localization and Mapping (SLAM) systems~\cite{strasdat2012local,engel2014lsd} and appearance-based robot relocalization~\cite{williams2011automatic}. For visual SLAM, state-of-the-art approaches~\cite{engel2014lsd} only handle a local window of recently added frames while the previous frames are marginalized out due to the limitation of computational complexity, resulting in the accumulation of state (position and orientation) error. LCD is introduced to identify places that have already been visited, thus creating an observation between history state and current state. The accumulated error can be effectively reduced based on this observation.

The most widely used LCD methods can be summarized as local feature based methods, which try to model image similarity based on hand crafted features. Most methods~\cite{cummins2008fab,angeli2008fast,labbe2013appearance,galvez2012bags,khan2015ibuild,mur2014fast} use Bag of Words (BOW) scheme to represent image since~\cite{sivic2003video}, which extracts feature points from an image and cluster them into different centroids called visual words. A histogram of appeared visual words is consequently used to represent the image. The similarity of image pairs is computed based on the difference of the visual words histograms. One well-known drawback of BOW is the perceptual aliasing introduced in cluster step if two dissimilar features are clustered into the same visual word. The performance of clustering depends on the quality of a previously~\cite{cummins2008fab} or online~\cite{angeli2008fast} trained dictionary.

Conventional methods~\cite{cummins2008fab,angeli2008fast,labbe2013appearance} using real-valued features like SIFT~\cite{lowe2004distinctive} or SURF~\cite{bay2006surf} suffer from high computational complexity in feature extraction and feature classification. To deal with this problem, recent methods like BOBW~\cite{galvez2012bags}, IBuILD~\cite{khan2015ibuild}, ORBSLAM~\cite{mur2014fast} have proposed to use efficient binary features like ORB~\cite{rublee2011orb} or BRISK~\cite{leutenegger2011brisk}. While binary feature based LCD methods can run at real time, the accuracy (typically measured by precision and recall metric~\cite{angeli2008fast}) of these methods is not satisfying.

In this paper, MILD: Multi-Index hashing for appearance based Loop closure Detection is proposed as an appearance based LCD approach exploiting the efficiency of binary features. Instead of using BOW representation widely adopted by previous methods, image similarity is measured based on direct feature matching without introducing additional computational complexity with the aid of Multi-Index Hashing (MIH)~\cite{greene1994multi}. Contributions of this paper include:
\begin{itemize}
  \item We propose a novel LCD system based on Multi-Index hashing (MILD). In particular, we do not explicitly find the exact nearest neighbor of each feature or use BOW representation for images. Instead, MIH is used to approximate the image similarity measurement, so that redundant computations between dissimilar features can be avoided.
  
  \item  The approximated image similarity measurement based on MIH is analyzed from a probabilistic perspective, which effectively reveals the trade-off between the accuracy and complexity in MILD, ensuring the superiority of MILD in high accuracy and low complexity compared with state-of-the-art algorithms.
      
  \item The detection of multiple loop closures is enabled in MILD, while most of the previous works~\cite{angeli2008fast,labbe2013appearance,zhang2011borf} assume that loop closure only occurs once in the candidate dataset for each query image.

\end{itemize}

\begin{figure*}[t]
\begin{minipage}[b]{1.0\linewidth}
  \centering
  \centerline{\epsfig{figure=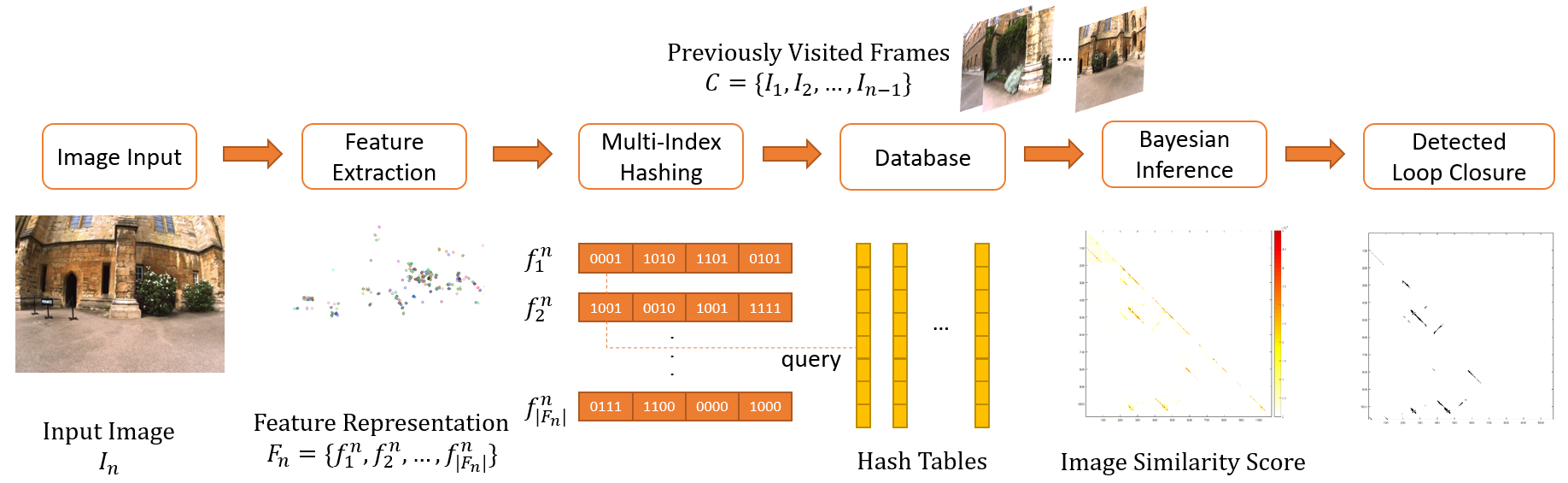,width=17cm}}
\end{minipage}
\caption{Overview of the proposed MILD: Multi-Index Hashing for Loop closure Detection.}
\label{fig:flowchart}
\end{figure*}

\section{RELATED WORK}

In this work, we focus on LCD by local image feature. Approaches such as global image descriptor~\cite{oliva2001modeling,kosecka2003qualitative} or exploiting illumination invariant components to improve image similarity measurement under different lighting conditions~\cite{maddern2014illumination} are not discussed, but can be combined for a more robust LCD system.

The accuracy of binary feature based LCD methods is not satisfying, the authors in~\cite{muja2012fast,lynen2014placeless} investigate this problem and find that binary features are not straightforward to cluster using existing nearest neighbor search methods, due to the high dimensionality and the nature of the binary descriptor space. To overcome this deficiency,~\cite{lynen2014placeless} projects binary features into a real-valued vector space and implements nearest neighbor search in this space.

An alternative way for LCD is direct feature match as proposed by~\cite{zhang2011borf,shahbazi2011application}. Instead of using BOW representation,~\cite{zhang2011borf} proposes to use raw features to represent an image directly (BoRF), which significantly improves the recall performance. \cite{shahbazi2011application} adopts Locality Sensitive Hashing (LSH) for fast approximate nearest neighbor search based on the SIFT feature. These methods suffer from high computational complexity and cannot scale well with the increase of candidate images.

We address this problem by Multi-Index Hashing (MIH) proposed by~\cite{greene1994multi} to hash long binary codes for fast information retrieval. Recently~\cite{norouzi2014fast} uses MIH for exact nearest neighbor search and tries to find the optimal substring length given the database size, code length and search radius to minimize the upper bound of the search cost. Experiments show that search cost grows rapidly with the increase of search radius.

As a method of nearest neighbor search, MIH has already been used in different applications like image relocalization~\cite{feng2016fast} and image search~\cite{cai2014scalable}.~\cite{feng2016fast} follows the same procedure in~\cite{norouzi2014fast} and complains about the inefficiency of MIH in finding the exact nearest neighbor for each feature. While~\cite{cai2014scalable} only explores the use of partial binary descriptors created in MIH as direct codebook indices, and follows a traditional BOW method to measure image similarity.

On the contrary to the previous methods, we do not explicitly find the exact nearest neighbor of each feature or use BOW representation for images. Instead, MIH is used to approximate the image similarity function proposed in~\cite{zheng2014coupled}. The accuracy and efficiency of such approximation are analyzed from a probabilistic perspective.

\section{MILD: Multi-Index Hashing for \\Loop closure Detection}

The framework of MILD is shown in Fig.~\ref{fig:flowchart}, where the MILD can be divided into two stages: the first step aims to calculate the similarity between current image $I_n$ and candidate set $C$ that are constructed by all the previous images $C=\{I_1, I_2, \cdots, I_{n-1}\}$. We denote \(F_k = \{f^k_1, f^k_2, \ldots, f^k_{|F_k|}\}\) as the binary local feature set to represent an image $I_k$, where $|F_k|$ stands for the number of features. Here the ORB feature [13] is used due to the computation efficiency and rotation invariance, with the descriptor be a 256 bit binary sequence. Given the image similarity, a Bayesian filter is applied to calculate the probability of loop closure for each candidate.

% which can be formulated as a binary segmentation problem and Markov Random Field is used to inference loop closure events \cite{anati2009constructing}.
\subsection{Image Similarity Measurement}

We define the similarity of image pair ($I_p$, $I_q$) as
\begin{equation}
\Phi(I_p,I_q) = \frac{ \overset{|F_p|}{\underset{i=1}{\sum}}  \overset{|F_q|}{\underset{j=1}{\sum}} \phi(f_i^p, f_j^q)}{|F_p||F_q|},
\label{Eqn:similarity}
\end{equation}
where $\phi(f_i^p,f_j^q)$ refers to binary feature similarity \cite{zheng2014coupled}, i.e.,
\begin{align}
\phi(f_i^p, f_j^q) =
\begin{cases}
exp(-d^2/\sigma^2), & d \leq d_0 \\
0,	& d > d_0.
\end{cases}
\end{align}
Here $d$ denotes Hamming distance between binary features $f_i^p$ and  $f_j^q$, $\sigma$ is the weighting parameter, and $d_0$ is the pre-defined Hamming distance threshold.

A straightforward way to calculate the image similarity is linear search for all the candidates in $C$. However, the computational cost may be unbearable for large datasets. Given the fact that the number of repeating or highly-similar features is limited between current image and previous images, implying that the valid similarity measurements are highly sparse, we propose to use Multi-Index Hashing (MIH) to avoid invalid computations, since MIH is capable in distinguishing similar features. More analysis is provided in Section~\ref{Sec_analysis}.

\begin{figure}[t]
\begin{minipage}[b]{1.0\linewidth}
  \centering
  \centerline{\epsfig{figure=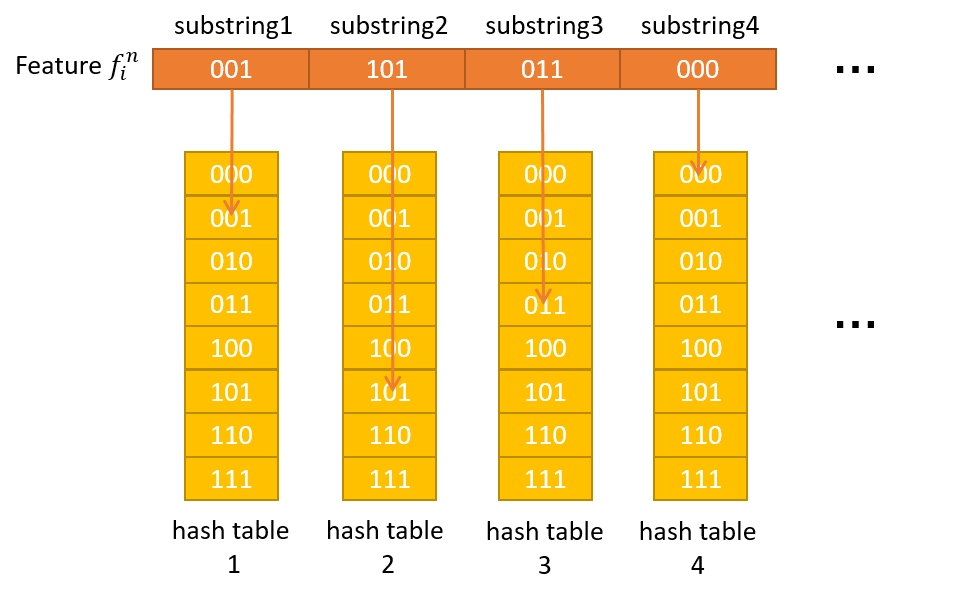,width=8.5cm}}
\end{minipage}
\caption{Framework of MIH. Binary feature $f_i^n$ is divided into $m$ disjoint substrings. $k$-{th} substring is the hash index of the $k$-{th} hash table. Image index $n$ and feature index $i$ are stored in corresponding $m$ entries as reference for feature $f_i^n$.}
\label{fig:MIH}
\end{figure}

As illustrated in Fig.~\ref{fig:MIH}, in MIH, a long binary feature is hashed $m$ times based on its $m$ disjoint substrings. More precisely, if the Hamming distance of two features is smaller than $L$, each feature is divided into $m$ disjoint substrings, then at least in 1 substring the Hamming distance of two features will be smaller than $L/m$ \cite{norouzi2014fast}, implying that for two features with small Hamming distance, the probability that they fall into the same entry in at least one hash table will be close to 1. Then, the image similarity measurement in Eqn. (\ref{Eqn:similarity}) can be approximated using MIH, where the database is constructed online based on the candidate set, and the image similarity is measured during the query stage. In practice, database construction and query are implemented with MIH simultaneously.
\begin{itemize}
  \item Database construction: For every input image ${I_k}$ and its feature set $F_k$, all features are hashed into the $m$ hash tables by separating each feature into $m$ substrings $\{h^k_{i,1}, \cdots, h^k_{i,m}\}$, where $h^k_{i,t}$ is the hash index of $t$-{th} hash table.
  \item Query: For the newly arrived query image $I_n$ and its binary feature set $F_n$, the similarity between $I_n$ and candidates $\{I_k, k = 1, \cdots, n-1\}$ is initialized as $0$. Let $\Omega_i^n$  be the collection of features that falls into the same entry with the feature $f_i^n$, then $\Phi(I_n, I_k)$ in Eqn. (\ref{Eqn:similarity}) can be approximated by
\begin{equation}
\Phi'(I_n,I_k) = \frac{\overset{|F_n|}{\underset{{i=1}}{\sum}} \underset{f_j^k \in \Omega_i^n}{\sum} \phi(f_i^n,f_j^k)}{|F_n||F_k|}.
\label{Eqn:approximate}
\end{equation}
\end{itemize}
Examining Eqn. (\ref{Eqn:approximate}), $\Phi'(I_n, I_k)$ can be calculated by 1 pass traverse of features in $F_n$ during the hashing process. $\Omega_i^n$ is a subset of $F_n$. The probability of $f_j^k, k \in [1, n-1], j \in [1, |F_k|]$ that falls into $\Omega_i^n$ (denoted as the recall probability) is related to the Hamming distance $d$ between $f_j^k$ and $f_i^n$, and the
number of hashing tables $m$ in MIH. The detailed analysis of the approximation error between $\Phi'(I_n, I_k)$ and $\Phi(I_n, I_k)$ is provided in Section \ref{Sec_analysis}.

\subsection{Bayesian Inference}
Bayesian inference is used to select true loop closure based on image similarity measurement and temporal coherency of camera movement \cite{angeli2008fast}. To enable the detection of multiple loop closures, we propose to extend the random variable representing loop closure hypotheses at time $t$ (denoted as $S_t$) to be a binary random variable $S_t^i$, where $S_t^i = 1$ is the event that current image $I_t$ closes the loop with the past image $I_i$. In this way, the time evolution model is formulated as
\begin{equation}
\left\{
\begin{array}{l}
P(S_t^i = 1 |S_t^{i-2:i+2}=1) = 0.9 \\
P(S_t^i = 1 |S_t^{i-2:i+2}=0) = 0.1 \\
P(S_t^i = 0 |S_t^{i-2:i+2}=1) = 0.1 \\
P(S_t^i = 0 |S_t^{i-2:i+2}=0) = 0.9
\end{array},\right.
\end{equation}
where $S_t^{i-2:i+2} \overset{\triangle}{=} \max \{S_t^j, j\in[i-2:i+2]\}$. Thus, the belief can be computed as
\begin{align}
B(k) = \sum_{j=0}^{1}P(S_t^i=k | S_t^{i-2:i+2}=j)P(S_t^{i-2:i+2} = j).
\end{align}

Recall that the image similarity measurement $\Phi'(I_i,I_t)$ is given by Eqn.~\ref{Eqn:approximate}, the likelihood is computed as~\cite{angeli2008fast}
\begin{align}
L(S_t^i = 1) =
\begin{cases}
\frac{\Phi'(I_t,I_i)-\sigma}{\mu}, & \Phi'(I_t,I_i) \leq \mu + \sigma, \\
1,	& \text{otherwise},
\end{cases}
\end{align}
where $\mu$ and $\sigma$ are the mean and standard deviation of sequence $\Phi'(I_t, I_i), i = 1, \cdots, t-1$. Finally, the loop closure probability $P(S^i_t = 1)$ given all the previous similarity measurements can be computed as
\begin{align}
P(S_t^i = 1)= \frac{L(S_t^i=1) \times B(1)}{\sum_{k=0}^{1}[L(S_t^i=k) \times B(k)]},
\end{align}
where $L(S^i_t = 0)$ is defined as a fixed value to normalize the output loop closure probability. The candidates whose loop closure probability is larger than the threshold $P_0$ will be the detected loop closures.

\subsection{Analysis of MIH}
\label{Sec_analysis}

\begin{figure}[htbp]
\begin{minipage}[htbp]{1.0\linewidth}
  \centering
  \centerline{\epsfig{figure=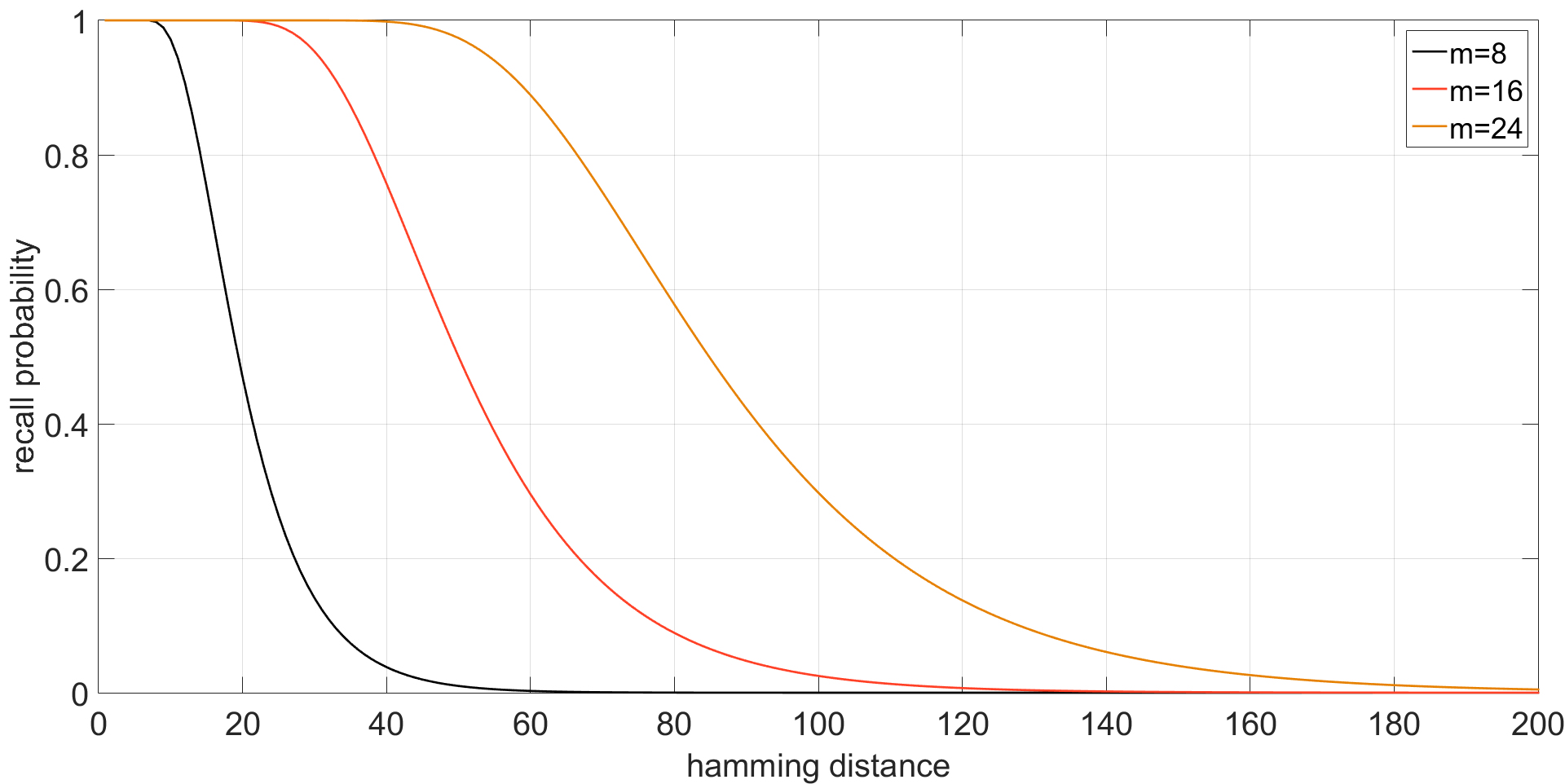,width=8cm}}
\end{minipage}
\caption{The effect of $d$ and $m$ on recall probability.}
\label{fig:recallProbability}
\end{figure}

Suppose the binary feature is divided into $m$ disjoint substrings, the probability that a feature pair with Hamming distance $d$ falls into the same entry in at least one of the $m$ hash tables is denoted as the recall probability $P_{recall}(m,d)$. This is equivalent to the case that $d$ independent balls are thrown into $m$ bins randomly, where the probability of at least one bin has no ball under the assumption of uniform distribution of Hamming errors is a solved problem~\cite{graham1994concrete}:
\begin{align}
P_{recall}(m,d) = 1 - \frac{m!\Theta(d,m)}{m^d}.
\end{align}
Here $\Theta(d, m)$ is the Stirling partition number~\cite{graham1994concrete}. Fig.~\ref{fig:recallProbability} shows the recall probability changes along Hamming distance $d$, as well as the influence of $m$ on the recall probability. As we expected, a larger $d$ yields a smaller recall probability, while a larger $m$ tends to make the decreasing curve of recall probability more gradual.

In LCD, for each feature in the query image, features describing the same place in $C$ are referred as inliers and the others are outliers. Then the computational complexity of $\Phi'$ (denoted as $E(m)$) is proportional to the average probability of outliers falling into $\Omega_i^n$. The accuracy of $\Phi'$ (denoted as $R(m)$) can be modeled as the average probability of inliers falling into $\Omega_i^n$. The unavoidable computations of similarity calculation for inliers are discarded in $E(m)$. Using the statistics of the distance distribution for inliers and outliers of ORB feature~\cite{rublee2011orb}, the Hamming distances of outliers and inliers can be modeled as Gaussian distribution $P_{o}(d) \in N(128,20)$ and $P_{i}(d) \in N(32,10)$, respectively. Based on this approximation, the accuracy and complexity can be calculated as
\begin{equation}
\begin{array}{cl}
R(m) &= \overset{L}{\underset{d=0}{\sum}} [P_{recall}(m, d)\times P_i(d)],	\\
E(m) &= \overset{L}{\underset{d=0}{\sum}} [P_{recall}(m, d)\times P_o(d)].
\end{array}
\label{Eqn:R_E}
\end{equation}

\begin{figure}[htbp]
\begin{minipage}[b]{1.0\linewidth}
  \centering
  \centerline{\epsfig{figure=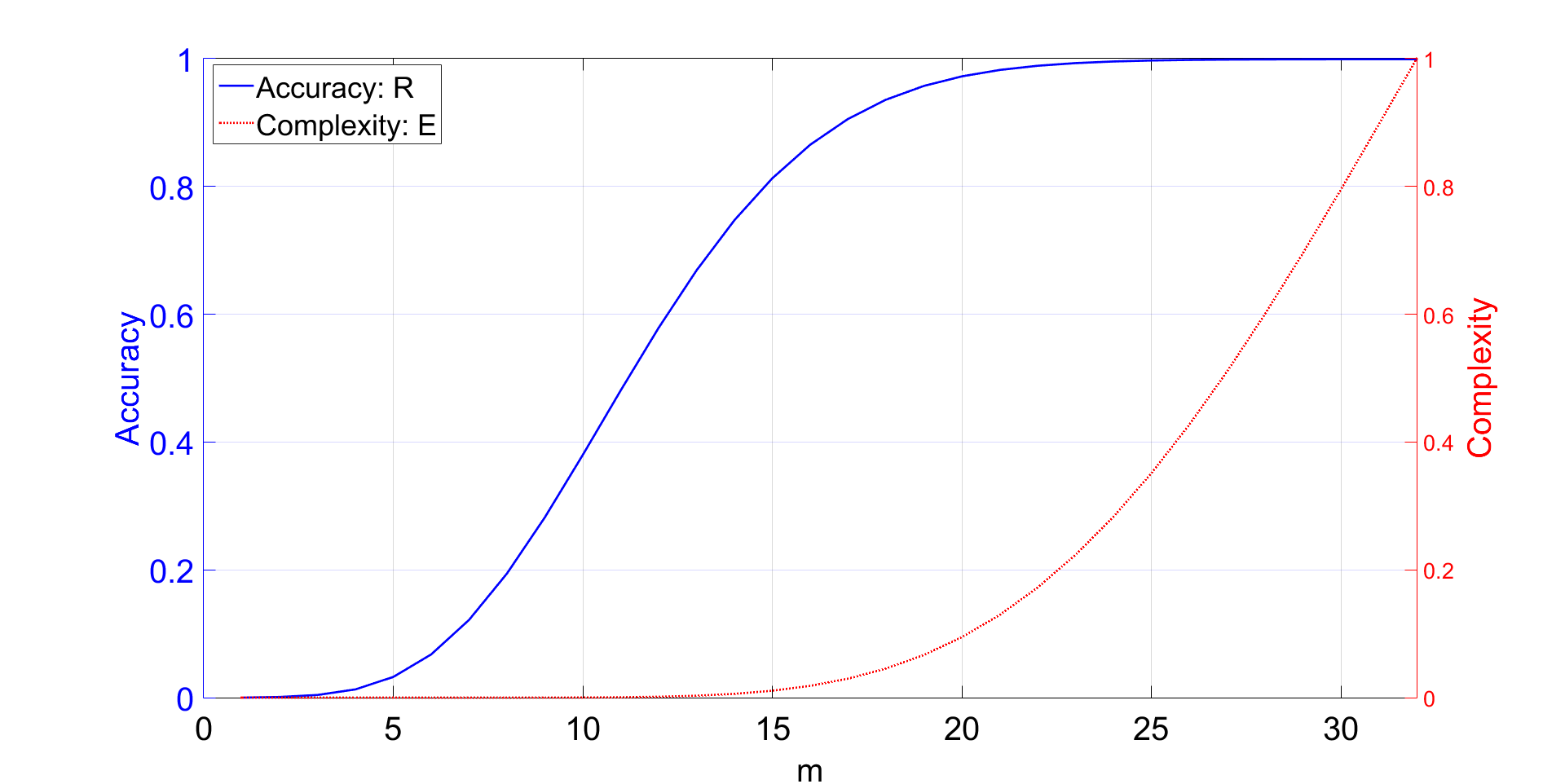,width=8cm}}
\end{minipage}
\caption{The effect of $m$ on accuracy $R$ and complexity $E$.}
\label{fig:recallComplexity}
\end{figure}

Given Eqn. (\ref{Eqn:R_E}), the influence of different $m$ on the trade-off between accuracy and complexity of MILD is further presented in Fig.~\ref{fig:recallComplexity}. A higher $R$ indicates that the approximation error between $\Phi'(I_n,I_k)$ and $\Phi(I_n,I_k)$ is smaller, yielding higher accuracy of MILD. While a lower $E$ indicates more efficiency. Although $R$ and $E$ grow monotonously with $m$, there exists an interval of $m$ to achieve good balance of high accuracy and low complexity. An appropriate $m$ can be chosen for different applications regarding different bias on accuracy and complexity. For example, in MILD, $m=16$ guarantees relatively high accuracy and very low computational cost. Experiments show that MILD enables loop closure detection within 15 ms for a database containing more than 1000 images, which is efficient enough for real-time LCD system.

\begin{figure*} \centering
\subfigure[Image Similarity Score] { \label{fig:a}
\includegraphics[height=4.8cm]{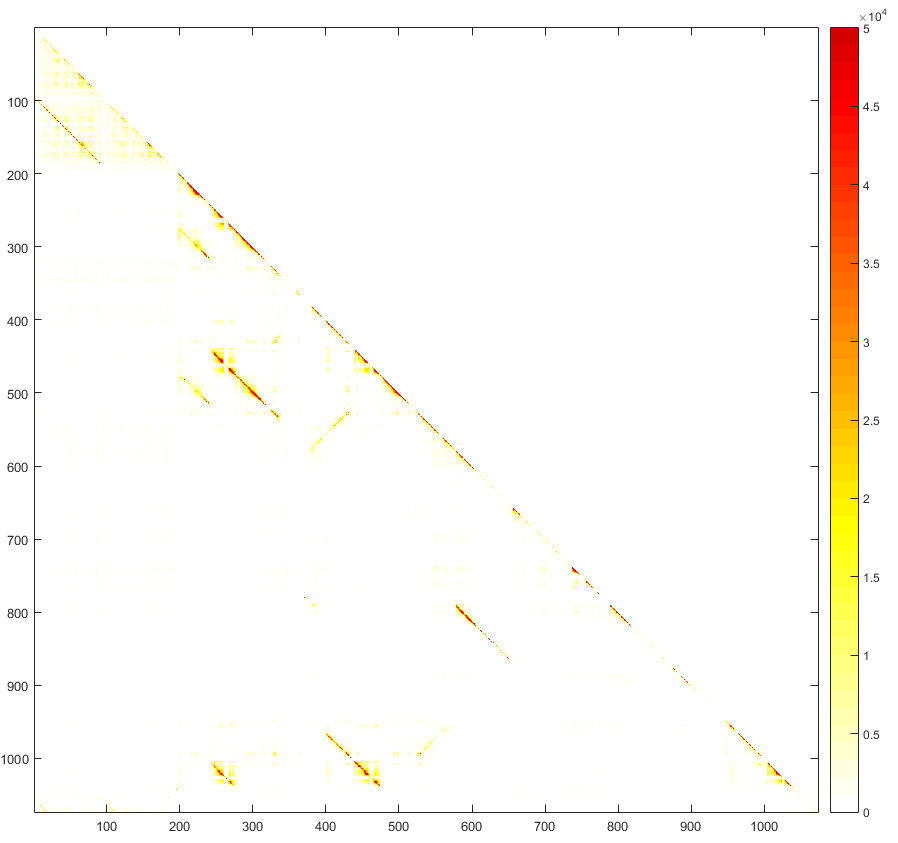} }
\subfigure[Detected Loop Closure] { \label{fig:b}
\includegraphics[height=4.8cm]{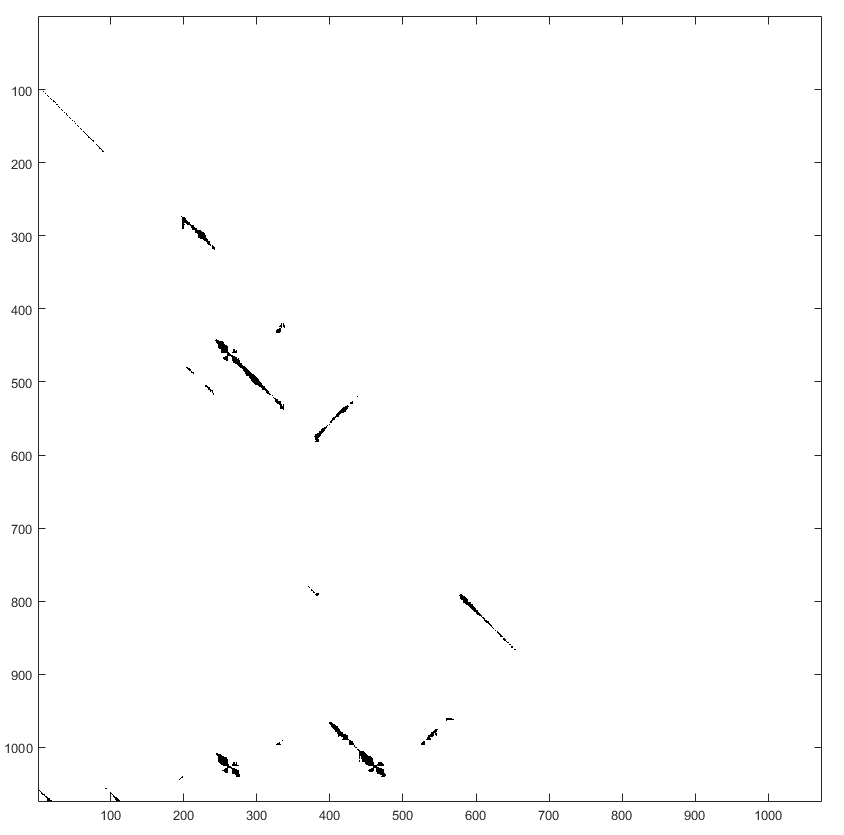} }
\subfigure[Ground Truth] { \label{fig:c}
\includegraphics[height=4.8cm]{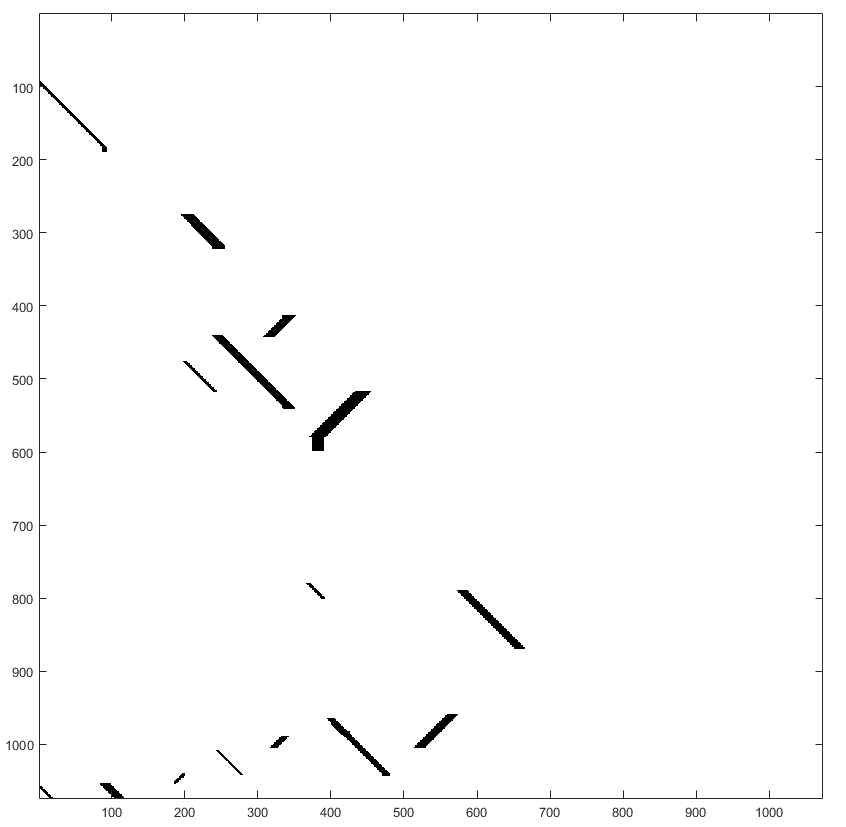} }
\caption{ Experiments on NewCollege Dataset~\cite{cummins2008fab}. The coordinate of each pixel $(i,j)$ represents the index for candidate image and query image respectively.}
\label{fig:NewCollege}
\end{figure*}

\vspace{-0.3cm}
\section{EXPERIMENTS AND DISCUSSIONS}
To evaluate the performance of MILD, we conduct extensive experiments on different datasets\footnote{NewCollege~\cite{cummins2008fab} contains 1073 images of size $1280\times 480$. CityCentre~\cite{cummins2008fab} contains 1237 images of size $1280\times 480$. Lip6Indoor~\cite{angeli2008fast} has 388 images of size $192\times 240$. Lip6Outdoor~\cite{angeli2008fast} has 1063 images of size $192\times 240$.} and compare with state-of-the-art methods: Angeli~\cite{angeli2008fast}, RTABMAP~\cite{labbe2013appearance} and BOWP~\cite{kejriwal2016high} which are based on SIFT/SURF feature, as well as BOBW~\cite{galvez2012bags} and IBuILD~\cite{khan2015ibuild} that use binary feature\footnote{All the Experiments are implemented on an Intel-core i7 @ 2.3 GHz processor with 8 GB RAM. Only one core is used to compare the computational efficiency of MILD with other algorithms. In MILD, 800 ORB features are extracted for each image. Feature descriptor is divided into 16 substrings with 16 bits each. The feature Hamming distance threshold $d_0=60$, and the loop closure probability threshold $P_0=0.7$.}. The implementation of MILD will be publicly available online.

\subsection{Subjective Analysis}
\label{Sec:subjective}

For a better understanding of MILD, we particularly show intermediate results of MILD on NewCollege dataset \cite{cummins2008fab} in Fig. \ref{fig:NewCollege}, where the approximated image similarity measurement using MIH is illustrated in Fig.~\ref{fig:a}. Given the image similarity measurement, Bayesian inference is employed to select loop closures among candidates, as shown in Fig.~\ref{fig:b}. Compared with the ground truth of loop closures (Fig.~\ref{fig:c}), the proposed MILD works effectively, as reflected by the fact that image similarity score in Fig.~\ref{fig:a} is high when image pair $(i,j)$ is a true loop closure, and the detected loop closures in Fig.~\ref{fig:b} highly resemble ground truth.

%{\small{\begin{table}[htbp]
%\begin{center}
%\caption{Public Dataset} \label{tab:dataset}
%\newcommand{\tabincell}[2]{\begin{tabular}{@{}#1@{}}#2\end{tabular}}
%\begin{tabular}{|c|c|c|c|c|}
%  \hline
%  	& \tabincell{c}{Image \\ Size} & \tabincell{c}{Image \\ Num} & \tabincell{c}{$T_1$\\ (ms)} & \tabincell{c}{ $T_2$\\ (ms)}
%  \\
%  \hline
%  \tabincell{c}{NewCollege~\cite{cummins2008fab}} & 1280x480 & 1073 & 35 & 11.5
%  \\
%  \hline
%
%  \tabincell{c}{CityCentre~\cite{cummins2008fab}} & 1280x480 & 1237 & 35 & 12
%  \\
%  \hline
%
%  \tabincell{c}{Lip6Indoor~\cite{angeli2008fast}} & 192x240 & 388 & 12 & 4
%  \\
%  \hline
%
%  \tabincell{c}{Lip6Outdoor~\cite{angeli2008fast}} & 192x240 & 1063 & 12 & 9
%  \\
%  \hline
%\end{tabular}
%\end{center}
%\end{table}}}
 
 \subsection{Objective Evaluation}
 \label{Sec:objective}
The quantitative comparisons regarding accuracy (recall rate at precision equals to 100\%) and complexity on different datasets are presented in~Table \ref{tab:comparison}, where the performance of concerned methods are collected directly from the reference papers. Examining Table \ref{tab:comparison}, we have following observations:

\begin{itemize}
  \item Angeli~\cite{angeli2008fast}, RTABMAP~\cite{labbe2013appearance} and BOWP~\cite{kejriwal2016high} require hundreds of milliseconds for LCD per image, due to the using of SIFT/SURF feature. Although RTABMAP yields the best recall, it takes 700ms per frame on average to process one query image, which is around \textbf{twenty times slower than MILD}.
  \item BOBW~\cite{galvez2012bags} and IBuILD~\cite{khan2015ibuild} that use binary feature can be implemented in real-time, but suffering from low accuracy, which is around \textbf{50\% lower than MILD}.
  \item On the contrary, although we do not assume single loop closure in the inference stage, which potentially introduces more outliers, MILD still achieves competitive performance in both accuracy and complexity, i.e., \textbf{the accuracy is comparable to SIFT/SURF feature based methods, and can be successfully implemented in realtime.}
\end{itemize}

\begin{table}[htbp]
\begin{center}
\caption{Comparisons with state-of-the-art algorithms} \label{tab:comparison}
\newcommand{\tabincell}[2]{\begin{tabular}{@{}#1@{}}#2\end{tabular}}
\begin{tabular}{|c|c|c|c|c|}
  \hline
  	& \tabincell{c}{City \\ Centre} & \tabincell{c}{New \\ College} & \tabincell{c}{Lip6 \\ Indoor} & \tabincell{c}{Lip6 \\ Outdoor}
  \\
  \hline
  \multirow{2}{*}{Angeli~\cite{angeli2008fast}} & - & - & 80\% & 71\% \\ \cline{2-5}
  						  & - & - & 460ms & 753ms
  \\
  \hline
  \multirow{2}{*}{RTABMAP~\cite{labbe2013appearance}} & 81\% & \textbf{89}\% & \textbf{98}\% & \textbf{95}\% \\ \cline{2-5}
  						  & 700ms & 700ms & 100ms & 400ms
  \\
  \hline
  \multirow{2}{*}{BOBP~\cite{kejriwal2016high}} & \textbf{86}\% & 77\% & 92\% & 94\% \\ \cline{2-5}
  						  & 441ms & 393ms & 69ms & 120ms
  \\
  \hline
  \multirow{2}{*}{BOBW~\cite{galvez2012bags}} & 30.6\% & 55.9\% & - & - \\ \cline{2-5}
  						  & 20ms & 20ms & - & -
  \\
  \hline
  \multirow{2}{*}{IBuILD~\cite{khan2015ibuild}} & 38\% & - & 41.9\% & 25.5\% \\ \cline{2-5}
  						  & - & - & - & -
  \\
  \hline
  \multirow{2}{*}{MILD} & 83\% & 87.3\%  & 94.5\% & 93.4\% \\ \cline{2-5}
  						  & 36ms & 35ms & 7ms & 9ms
  \\
  \hline
\end{tabular}
\end{center}
\end{table}

For memory requirement, MIH takes 32 bytes to store feature descriptors and 4 bytes to store its corresponding image index and feature index in each hash table per feature. The only fixed overhead of MILD is \(2^l\) pointers for each hash table, where $l$ is the substring length. In our experiments, $l=16$ and there are $16$ hash tables in total. For example, the minimum memory required for NewCollege dataset is $84$ MB, which is acceptable for modern mobile devices.

%\subsection{Sparse Match}
%
%Based on the analysis of trade-off between efficiency and accuracy in MIH, a light version of MILD -- sparse match -- is further proposed for binary feature matching on different images. For image matching, overhead of database construction must be small and accuracy of the match is more important than LCD. Thus, in sparse match, the hash table number is set as $m=24$, which yields a smaller hash table and a higher recall than $m= 16$ used in MILD.
%
%To verify the complexity and accuracy of sparse match, we select 1000 consecutive image pairs in NewCollege Dataset as test-data. Exhaustive search implemented in OpenCV \cite{bradski2000opencv} serves as benchmark. The matches whose Hamming distance is smaller than 60 are considered as valid. We found sparse match can find 90\% valid matches while being 10 times faster than exhaustive search methods. The implementation of MILD and sparse match will be publicly available online.

\section{CONCLUSIONS AND FUTURE WORK}
While MIH has shown large potential in exactly nearest neighbor search recently \cite{norouzi2014fast}, we extend its application in approximately nearest neighbor search and propose a novel Multi-Index Hashing scheme for Loop closure Detection problem (MILD). Theoretical analysis successfully reveals the trade-off between accuracy and efficiency of MIH in image similarity measurement. Experiments on public datasets show that MILD achieves competitive performance regarding high accuracy and low complexity, compared with state-of-the-art LCD approaches.

In our work, the uniform distribution of binary codes is assumed, but in practice many features fall into the same entry in the hashing process, such entries are discarded for efficiency consideration. It would be interesting to consider prior knowledge on non-uniform distribution of different features for improving MILD.

\bibliographystyle{IEEEbib}
\small{\bibliography{MILD}}
%{\small{}}

\end{document}